%% file: main.tex
\documentclass[fleqn,10pt,dvipsnames]{wlscirep}
\usepackage[utf8]{inputenc}
\usepackage[T1]{fontenc}
\usepackage{siunitx}
\usepackage{multirow}
\usepackage{tabularx}
\usepackage{framed}
\usepackage[capitalise,noabbrev]{cleveref}

\DeclareSIUnit{\million}{\text{million}}
\DeclareSIUnit{\billion}{\text{billion}}
\DeclareSIUnit{\tokens}{\text{tokens}}
\DeclareSIUnit{\year}{\text{year}}
\DeclareSIUnit{\years}{\text{years}}
\DeclareSIUnit{\hours}{\text{hours}}
\DeclareSIUnit{\days}{\text{days}}
\DeclareSIUnit{\weeks}{\text{weeks}}

\newcommand{\hlsection}[1]{\textcolor{ForestGreen}{\textbf{\textsc{\MakeLowercase{#1}}}}}
\newcommand{\hlsubsection}[1]{\textbf{\textsc{\MakeLowercase{#1}}}}

\title{Clinical Language Understanding Evaluation (CLUE)}

\author[1,*]{Travis R. Goodwin}
\author[1]{Dina Demner-Fushman}
\affil[1]{U.S. National Library of Medicine, Intramural Research Program, Bethesda, 20894, USA}
\affil[*]{corresponding author(s): Travis R. Goodwin}

\begin{abstract}
Clinical language processing has received a lot of attention in recent years, resulting in new models or methods for disease phenotyping, mortality prediction, and other tasks. 
Unfortunately, many of these approaches are tested under different experimental settings (e.g., data sources, training and testing splits, metrics, evaluation criteria, etc.) making it difficult to compare approaches and determine state-of-the-art. 
To address these issues and facilitate reproducibility and comparison, we present the Clinical Language Understanding Evaluation (CLUE) benchmark with a set of four clinical language understanding tasks, standard training, development, validation and testing sets derived from MIMIC data, as well as a software toolkit.
It is our hope that these data will enable direct comparison between approaches, improve reproducibility, and reduce the barrier-to-entry for developing novel models or methods for these clinical language understanding tasks.
\end{abstract}

\begin{document}

\flushbottom
\maketitle

\thispagestyle{empty}


\section*{Background \& Summary}

In the general domain, there has been remarkable progress on many natural language processing tasks thanks to the advent of transformer-based models such as ELmo\cite{peters-etal-2018-deep}, OpenAI GPT (1-3)\cite{radford2018improving,radford2019language,brown2020language}, BERT\cite{devlin-etal-2019-bert}, and T5\cite{raffel2020exploring}.
These models exemplify the promise of self-supervised pre-training from massive unlabelled text datasets allowing for transfer of generalized knowledge to new tasks of interest (i.e., fine-tuning) with substantially reduced training requirements.
The rapid advances in transformer-based model development has greatly benefited from the existence of standard benchmark corpora, such as the General Language Understanding Evaluation\cite{wang2019glue} (GLUE) benchmark -- a collection of nine language understanding tasks build on existing datasets covering a range of dataset sizes, genres, and task difficulties, all with standardized training, validation, and testing splits.
The existence of freely-available standard benchmark collections allow researchers to directly compare the performance of different natural language processing approaches, quantify improvements, and determine the state-of-the-art.

By contrast, the clinical domain greatly suffers from data fragmentation and lack of resource sharing.
For this reason, methodological contributions in clinical natural language processing are often difficult or impossible to evaluate against related work.
Determining the state-of-the-art for various clinical language processing tasks is equally difficult.
Even when two approaches are evaluated on the same dataset, they often rely on different experimental settings, e.g., training, validation, and testing splits; evaluation methods; and reported metrics hindering direct comparison.
This often results in duplication of research effort and significantly increases the barriers to selecting an optimal approach for eventual implementation.

In this paper, we propose a standard, freely-available benchmark collection for Clinical Language Understanding Evaluation (CLUE).
We consider six clinical language understanding tasks based on disease staging, disease phenotyping, all-cause mortality prediction and remaining length of stay prediction.
For these six tasks, we use clinical records from MIMIC-III\cite{johnson2016mimic,pollard2016mimic}, a freely-accessible critical care database\cite{goldberger2000physionet}. 
All tasks are carefully split into standard stratified training, validation, calibration, and testing sets that preserve the distributions of potentially confounding demographic and admission information, thus reducing the likelihood of learning bias and over-fitting.


\section*{Methods}
The goal of CLUE is to provide a flexible and robust mechanism for evaluating any automatic method for clinical language understanding. 
To accomplish this, we were interested in ensuring CLUE was (1) accessible -- requiring minimal individual data usage agreements as possible, (2) flexible -- providing data in as close to original format as possible (i.e., with minimal de-identification artifacts), (3) meaningful -- covering tasks that have direct clinical impact, and (4) robust -- covering a range of tasks such that success on all tasks requires clinical language understanding rather than surface-level pattern recognition. 

\subsection*{Clinical Data Source}
A limited number of clinical document collections have been released to the community. 
The most well-known are: (1) the University of Pittsburgh BluLAB NLP repository, containing (in 2012) \num{93551} de-identified clinical reports for \num{17264} hospital visits; (2) the n2c2 (National NLP Clinical Challenges) research data sets; and (3) 
the MIMIC-III critical care database\cite{johnson2016mimic} containing \num{2082284} de-identified clinical notes for \num{46520} patients and \num{58976} hospital admissions.
Of these three, the Pittsburgh collection is no longer freely available.
Both the n2c2 and MIMIC-III datasets are freely available, however, the n2c2 datasets are fragmented into relatively small datasets with specialized cohorts for specific NLP challenges making them difficult to use as a benchmark for data-hungry deep learning methods. 
Moreover, MIMIC-III includes a wealth of time-stamped structured data about each patient including laboratory values, vital signs, demographics, chart information, and discharge billing diagnoses.
In CLUE we chose to base all tasks off of the MIMIC-III dataset so-as to (1) enable the use of CLUE with a single data usage agreement, (2) expand the number of documents we could label by taking advantage of structured data associated with clinical notes recorded on the same day, and (3) enable multi-task learning on a large general-purpose cohort of patients. 

\paragraph{Exclusion Criteria}
We excluded patients under the age of 15 (primarily neonates), and excluded hospital admissions with fewer than two days with clinical notes.

\subsection*{Clinical Task Selection}
We considered four main clinical language understanding tasks: 
(1) disease staging, 
(2) clinical phenotyping,
(3) mortality prediction, and 
(4) remaining length-of-stay prediction.

\subsubsection*{Disease Staging}
The ability to reliably recognizing the stage of a disease at different points in a patient's care is critical for assessing changes in severity of illness for specific populations or specific episodes of care. 
Moreover, disease staging is important for evaluating healthcare processes as well as outcomes. 
For this reason, we included three disease staging tasks, each with (a) well-defined clinical stages that can be directly measured from structured data based on clinical guidelines, (b) sufficient representation in MIMIC-III, and (3) different patterns of evidence (measurements, qualitative descriptions, etc.). 

\paragraph{Acute Kidney Injury}
Acute kidney injury (AKI) affects as many as \SI{20}{\percent} of all hospitalizations resulting in an estimated cost of \SI{10}[\$]{\billion} annually.\cite{silver2017cost,chertow2005acute}
AKI is associated with increased mortality, end-stage renal disease, and chronic kidney disease.\cite{chertow2005acute}
It has been shown that even small increases in serum creatinine are associated with long-term damage and increased mortality.\cite{chertow2005acute}
The 2011 Kidney Disease: IMproving Global Outcomes (KDIGO) Clinical Priactice GHuideline for Acute Kidney Injury (AKI) provides clear clinical criteria for staging AKI as well as evidence-based management recommendations for each AKI stage.\cite{kellum2012kidney}
\Cref{tab:kdigo} presents our criteria for determining AKI stages, based on the KDIGO criteria.\footnote{%
In the KDIGO criteria, Stage 3 can also be determined by the initiation of renal replacement therapy, present of Anuria for $\geq$ \SI{12}{\hours} or, in patients $<$ \SI{18}{\years}, a decrease in eGFR to $<$ \SI{35}{\milli\litre\per\minute} per \SI{1.73}{\meter\squared}.
These events are rare and difficult to cleanly detect in MIMIC-III, thus, for simplicity, we consider only the serum creatinine and urine output criteria when staging AKI.%
}
For example, KDIGO recommends non-invasive diagnostic workup at Stage 1, checking for changes in drug dosing at Stage 2, and to avoid subclavian catherization at Stage 3.
Thus, the ability to automatically identify the stage of AKI at different timepoints would enable earlier changes to management and could help prevent (further) deterioration of kidney function. 
Note: when assigning AKI labels for a note, we used the highest AKI stage evidenced by that patient's structured measures on or before the date of the note.
\input{floats/kdigo}

\paragraph{Pressure Injury}
The development of pressure injuries (i.e., pressure ulcers or bed sores) can lead to several complications, including sepsis, cellulitis, osteomyelitis, pain, and depression.\cite{brem2010high}
The mortality rate has been noted to be as high as \SI{60}{\percent} within \SI{1}{\year} of hospital discharge for older patients who develop a pressure ulcer during their stay.\cite{thomas1996hospital} 
The National Pressure Injury Advisory Panel recognizes four stages of pressure injuries.
The NPIAP descriptions of each stage, as well as the MIMIC-III chart values corresponding to those stages are provided in \cref{tab:npiap}.
While structured descriptions of pressure injuries are sometimes available in the patient's chart, the specific characteristics of pressure injuries are often documented in unstructured clinical narrative.
The ability to automatically asses the stage of pressure injury from clinical notes would enable earlier intervention and automatic processes for preventing further decomposition.
Note: when assigning labels for a note, we used the highest stage for any pressure injury recorded for that patient on or before the date of the note.
\input{floats/npiap}

\paragraph{Anemia}
{%
\sisetup{%
    range-phrase = --,%
    range-units = single,%
    per-mode = repeated-symbol
    }%
A substantial number of hospital patients with normal HgB on admission become anemic during the course of their hospitalization resulting in increased average length of stay by \SIrange{10}{88}{\percent}, hospital charges by \SIrange{6}{80}{\percent}, and risk of in-hospital mortality by \SIrange{51}{228}{\percent}, depending on anemia severity.\cite{koch2013hospital}%
\Cref{tab:who} defines the criteria used to assign anemia stages for each patient.
Automatically assessing the severity of anemia would enable to physicians to prevent further deterioation by switching to small volume phlebotomy tubes, minimizing blood loss from in-dwelling catheters, and reducing blood tests.\cite{chant2006anemia,harber2006highly}
Note: when assigning labels for a note, we used the lowest HgB levels recorded on the same calendar date of that note.
}%
\input{floats/who}

\subsubsection*{Computational Phenotyping}
The ability to identify patients with particular conditions is important for clinical quality measurement, health improvement and research. 
This process, known as computational phenotyping,\cite{hripcsak2012next} allows decision makers to identify and target patients for screening or specific interventions that have been shown particularly effective in similar populations.\cite{richesson2020electronic}
Automatically detecting a patient's computational phenotype can support interventional, observational, prospective, and retrospective studies. \cite{richesson2020electronic,alzoubi2019review}
Moreover, Denny et al. (2012)\cite{denny2012chapter} points out that phenotype identification is one task for which clinical notes are particularly useful, noting that this is often because salient observations in text documents such as pathology and radiology reports are often not also included in tabular data.

Using standard phenotype definitions is important to facilitate re-use, reproducability, and generalization. 
Consequently, as in mimic benchmark, other benchmark, elsevier benchmark, we definine phenotypes using semantically meaningful groups of discharge diagnostic codes (ICD-9).
Specifically, we rely on the hierarchical groupings of ICD-9 codes published by the Agency for Healthcare Research and Quality (AHRQ) and used in their Clinical Classification Software (CCS). 
CCS codes are groups of ICD-9 codes that correspond to specific diseases; these CCS codes are further grouped into a hierarchy based on organ systems and disease categories.
We consider the presence of any ICD-9 code in each of these CCS groupings as evidence for the corresponding clinical phenotype.

\subsubsection*{Mortality Prediction}
{%
\sisetup{%
    range-units = single,%
    list-final-separator = {, or },%
    list-pair-separator = { or },
    list-units = single%
    }%
Mortality is one of the primary outcomes of concern for hospitalized patients: the ability to accurately predict mortality could enable caregivers to optimize treatment and prioritize resources. 
We look at mortality in several settings:
\begin{itemize}[nosep]
    \item Short-term mortality prediction in which the goal is to predict whether a patient will die within \SIlist{24;48;72}{\hours}.
    \item Long-term mortality prediction in which the goal is to predict if the patient dies within \SIlist{10;30;90}{\days}, or within \SI{1}{\year}.
\end{itemize}
While mortality is often predicted using data obtained during the first \SIrange{24}{48}{\hours} of a patient's hospital admission, as a mechanism for testing clinical language understanding, we provide all mortality labels for each clinical note in a patient's stay, with each time window relative to the timestamp of that note. 
Thus, the mortality prediction task in CLUE can be viewed as a generalization of most \SIrange[range-phrase = --]{24}{48}{\hour} mortality prediction settings (where the traditional setting can be recovered by considering only notes within the first \SIrange{24}{48}{\hours}).
}

\subsubsection*{Length-of-Stay Prediction}
{
\sisetup{%
    list-units = single,%
    list-final-separator = {, or },%
    }%
Length-of-stay is an important metric for both patients and healthcare providers and is of particular importance in critical care in which length-of-stay is heavily correlated with mortality.
Planning bed availability administration.
In CLUE we provide remaining length-of-stay labels for windows of \SIlist{1;2;3;4;5;6;7;8}{\days}; as well as a single label for more than \SI{8}{\days} but less than \SI{2}{\weeks}. 
}

\subsection*{Data Randomization and Stratification}
To minimize the affect of demographic and other confounders, we split data into stratified 8:1:1:1 splits for training, validation, calibration, and testing, respectively.
Splits were stratified at the patient level to preserve the same distribution of demographic and admission information including the patient's age, sex, and race as well as their admitting ICU, source of admission (viz., clinic, physician, transfer, or other), type of admission (viz., elective, emergency, or urgent), Oxford Acute Severity of Illness Score\cite{johnson2013new} (OASIS), and type of insurance (viz., government, private, Medicaid, Medicare, or self pay).
We used an iterative stratification method based on Sechidis et al. (2011)\cite{sechidis2011stratification} and Szyma\'{n}ski and Kajdanowicz (2017)\cite{szymanski2017network} to stratify data.\cite{szymanski2017scikit}

\subsection*{Scoring}
As with GLUE\cite{wang2018glue} and Super GLUE\cite{wang2019superglue}, we provide an evaluation script to evaluate system performance on CLUE.
We assess disease staging as an ordinal regression task, relying on Mean Absolute Error (MAE) as the primary metric.
We cast length-of-stay prediction as a standard regression task (where the remaining length of stay is measured in days) using MAE as the primary metric.
Mortality prediction is cast as a multi-label binary classification task, using macro-average $F_1$-measure as the primary metric.
Finally, phenotyping is also cast as multi-label binary classification, measured with micro-average $F_1$.
When designing CLUE we wanted to also provide a sense of aggregate system performance over all tasks.
To do this, we compute the macro-average performance over all tasks.
Because staging and length-of-stay are evaluated in terms of error, for the purposes of aggregation, MAE scores are converted to Mean Average Accuracy (MAA), i.e., $MAA = 1 - MAE$.



\section*{Data Records}
The CLUE dataset is provided in several CSV files, corresponding to the training, validation, calibration, and testing splits for each of the tasks.
Each row in the CSV consists of the following fields:
\begin{enumerate}[nosep]
    \item the MIMIC-III hospital admission ID associated with the note(s); 
    \item the timestamp and date of the note(s);
    \item the free-text content of one or more clinical notes extracted on the same calendar day; and
    \item the task-specific label(s) for the associated timestamp and date
\end{enumerate}
For the staging tasks, labels are provided in the form of a single integer, indicating the stage of the disease (where zero indicates normal), for phenotyping, the labels are provided as a list of present phenotypes (a comma separated list of CCS codes), and for length-of-stay and mortality prediction, the labels are provided as multiple columns with the value of True or False for each length-of-stay or mortality window.
\Cref{fig:records} illustrates several example records, while \cref{fig:note} illustrates an example note.
Table X shows demographics for our training, development, calibration, and testing splits.
\input{floats/records}
\input{floats/note}

\subsection*{Dataset Toolkit}
In addition to the raw CSV files and evaluation script, we provide a python toolkit for working with CLUE.
The CLUE toolkit includes scripts for replicating the experiments reported in this paper, as well as dataset and metric descriptors for the HuggingFace's Transformers\cite{wolf-etal-2020-transformers} and Dataset\cite{wolf2020datasets} packages.
HuggingFace Datasets is a python library for easily sharing and accessing datasets and evaluation metrics for Natural Language Processing (NLP), with built-in interoperability with Numpy\cite{harris2020array}, Pandas\cite{mckinney2010data,reback2020pandas}, PyTorch\cite{paszke2019pytorch} and Tensorflow\cite{tensorflow2015whitepaper,tfds}.
Moreover, the HuggingFace integration allows for Transformer-based models such as BERT, T5, and BigBird to be easily used with CLUE.

\section*{Usage Notes}%
The original MIMIC-III database analyzed in this study and the resultant CLUE dataset are both available through PhysioNet.



\section*{Code availability}
The code to extract data and labels from MIMIC-III is available on Zenodo, the code to generate the CLUE datasets from these labels is also available on Zenodo and GitHub, the CLUE software toolkit is available separately on GitHub.

\bibliography{main,acl_anthology}

\section*{Acknowledgements}
This work was supported by the intramural research program at the U.S. National Library of Medicine, National Institutes of Health, and utilized the computational resources of the NIH HPC Biowulf cluster (\href{http://hpc.nih.gov}{http://hpc.nih.gov}).

\section*{Author contributions statement}
T.G. conceived and conducted the experiments. T.G. and D.F. analyzed the results.
All authors reviewed the manuscript.

\section*{Competing interests}
The authors declare no competing interests.

\end{document}

%% file: floats/kdigo.tex
\begin{table}[!hp]
    \centering
    \sisetup{%
    range-phrase = --,%
    range-units = single,%
    per-mode = repeated-symbol%
    }%
    \begin{tabularx}{\linewidth}{lX}
    \toprule
         AKI Stage & KDIGO Criteria \\
    \midrule
        1   &   \begin{itemize}[nosep, before=\leavevmode\vspace*{-1\baselineskip}, after=\leavevmode\vspace*{-1\baselineskip}, wide, labelwidth=!, labelindent=0pt]
                \item \SIrange{1.5}{1.9}{} $\times$ baseline serum creatinine 
                \item $\geq$ \SI{0.3}{\milli\gram\per\deci\litre} $\left(\geq \text{\SI{26.5}{\micro\mole\per\litre}}\right)$ increase in serum creatinine
                \item $<$ \SI{0.5}{\milli\litre\per\kilo\gram\per\hour} urine ouptut for \SIrange{6}{12}{\hours} 
                \end{itemize}
            \\ \\
            
        2   &   \begin{itemize}[nosep, before=\leavevmode\vspace*{-1\baselineskip}, after=\leavevmode\vspace*{-1\baselineskip}, wide, labelwidth=!, labelindent=0pt]
                \item \SIrange{2.0}{2.9}{} $\times$ baseline serum creatinine
                \item $<$ \SI{0.5}{\milli\litre\per\kilo\gram\per\hour} urine output for $\geq$ \SI{12}{\hours}
                \end{itemize}
            \\ \\
            
        3   &   \begin{itemize}[nosep, before=\leavevmode\vspace*{-1\baselineskip}, after=\leavevmode\vspace*{-1\baselineskip}, wide, labelwidth=!, labelindent=0pt]
                \item 3.0 $\times$ baseline serum creatinine
                \item $\geq$ \SI{4.0}{\milli\gram\per\deci\litre} $\left(\geq \text{\SI{353.6}{\micro\mole\per\litre}}\right)$ increase in serum creatinine
                \item  $<$ \SI{0.5}{\milli\litre\per\kilo\gram\per\hour} urine output for $\geq$ \SI{24}{\hours} 
                \end{itemize}
            \\

    \bottomrule
    \end{tabularx}
    \caption{KDIGO criteria for staging Acute Kidney Injury (AKI)}
    \label{tab:kdigo}
\end{table}

%% file: floats/npiap.tex
\begin{table}[!hp]
    \centering
    \begin{tabularx}{\linewidth}{lXX}
    \toprule
    Stage &  NPIAP Description & MIMIC-III Chart Values \\
    \midrule
         1 
             & \textbf{Non-blanchable erythema of intact skin:} 
Intact skin with a localized area of non-blanchable erythema, which may appear differently in darkly
pigmented skin. Presence of blanchable erythema or changes in sensation, temperature, or firmness
may precede visual changes. Color changes do not include purple or maroon discoloration; these
may indicate deep tissue pressure injury.  
             &  \begin{itemize}[nosep, before=\leavevmode\vspace*{-1\baselineskip}, after=\leavevmode\vspace*{-1\baselineskip}, wide, labelwidth=!, labelindent=0pt]
                \item Unable to assess; dressing not removed
                \item Red, Unbroken
                \item Red; unbroken
                \item Intact,Color Chg
                \item Unable to Stage
                \item Other/Remarks
                \item Deep Tiss Injury
                \item Deep tissue injury
                \end{itemize}
         \\
         \\
         
         2 
             & \textbf{Partial-thickness skin loss with exposed dermis:}
Partial-thickness loss of skin with exposed dermis. The wound bed is viable, pink or red, moist, and
may also present as an intact or ruptured serum-filled blister. Adipose (fat) is not visible and deeper
tissues are not visible. Granulation tissue, slough and eschar are not present. These injuries
commonly result from adverse microclimate and shear in the skin over the pelvis and shear in the
heel. This stage should not be used to describe moisture associated skin damage (MASD) including
incontinence associated dermatitis (IAD), intertriginous dermatitis (ITD), medical adhesive related
skin injury (MARSI), or traumatic wounds (skin tears, burns, abrasions).  
             &  \begin{itemize}[nosep, before=\leavevmode\vspace*{-1\baselineskip}, after=\leavevmode\vspace*{-1\baselineskip}, wide, labelwidth=!, labelindent=0pt]
                \item Part. Thickness
                \item Partial thickness skin loss through epidermis and/or dermis; ulcer may present as an abrasion, blister, or shallow crater
                \item Partial thickness skin loss through epidermis and/or dermis; ulcer may present as an abrasion, blister, or shallow crater
                \item Through Dermis
                \end{itemize}
         \\
         \\
         
         3 
             & \textbf{Full-thickness skin loss:}
Full-thickness loss of skin, in which adipose (fat) is visible in the ulcer and granulation tissue and
epibole (rolled wound edges) are often present. Slough and/or eschar may be visible. The depth of
tissue damage varies by anatomical location; areas of significant adiposity can develop deep
wounds. Undermining and tunneling may occur. Fascia, muscle, tendon, ligament, cartilage and/or
bone are not exposed. If slough or eschar obscures the extent of tissue loss this is an Unstageable
Pressure Injury.  
             &  \begin{itemize}[nosep, before=\leavevmode\vspace*{-1\baselineskip}, after=\leavevmode\vspace*{-1\baselineskip}, wide, labelwidth=!, labelindent=0pt]
                \item Full Thickness
                \item Full thickness skin loss that may extend down to underlying fascia; ulcer may have tunneling or undermining
                \item Unable to stage; wound is covered with eschar
                \end{itemize}
         \\
         \\
         
         4 
             & \textbf{Full-thickness skin and tissue loss:}
Full-thickness skin and tissue loss with exposed or directly palpable fascia, muscle, tendon,
ligament, cartilage or bone in the ulcer. Slough and/or eschar may be visible. Epibole (rolled edges),
undermining and/or tunneling often occur. Depth varies by anatomical location. If slough or eschar
obscures the extent of tissue loss this is an Unstageable Pressure Injury. 
             &  \begin{itemize}[nosep, before=\leavevmode\vspace*{-1\baselineskip}, after=\leavevmode\vspace*{-1\baselineskip}, wide, labelwidth=!, labelindent=0pt]
                \item Full thickness skin loss with damage to muscle, bone, or supporting structures; tunneling or undermining may be present
                \item Through Fascia
                \item To Bone
                \end{itemize}
         \\
         \bottomrule
    \end{tabularx}
    \caption{Pressure Injury stages as defined by the National Pressure Injury Advisory Panel (NPIAP) and their corresponding CareVue and MetaVision chart values in MIMIC-III}
    \label{tab:npiap}
\end{table}

%% file: floats/who.tex
\begin{table}[!hp]
    \sisetup{%
    range-phrase = --,%
    range-units = single,%
    per-mode = repeated-symbol%
    }
    \centering
    \begin{tabularx}{\linewidth}{lX}
    \toprule
    Stage & WHO Haemoglobin Criteria \\
    \midrule
    1 & \begin{itemize}[nosep, before=\leavevmode\vspace*{-1\baselineskip}, after=\leavevmode\vspace*{-1\baselineskip}, wide, labelwidth=!, labelindent=0pt]
        \item{} \SIrange{110}{119}{\gram\per\litre}, for non-pregnant women
        \item{} \SIrange{100}{109}{\gram\per\litre}, for pregnant women
        \item{} \SIrange{110}{129}{\gram\per\litre}, for men
        \end{itemize} \\ \\
        
    2 & \begin{itemize}[nosep, before=\leavevmode\vspace*{-1\baselineskip}, after=\leavevmode\vspace*{-1\baselineskip}, wide, labelwidth=!, labelindent=0pt]
        \item{} \SIrange{80}{109}{\gram\per\litre}, for non-pregnant women
        \item{} \SIrange{70}{99}{\gram\per\litre}, for pregnant women
        \item{} \SIrange{80}{109}{\gram\per\litre}, for men
        \end{itemize} \\ \\
        
    3 & \begin{itemize}[nosep, before=\leavevmode\vspace*{-1\baselineskip}, after=\leavevmode\vspace*{-1\baselineskip}, wide, labelwidth=!, labelindent=0pt]
        \item{} $<$ \SI{80}{\gram\per\litre}, for non-pregnant women
        \item{} $<$ \SI{70}{\gram\per\litre}, for pregnant women
        \item{} $<$ \SI{80}{\gram\per\litre}, for men
        \end{itemize}
        \\
    \bottomrule
    \end{tabularx}
    \caption{World Health Organization (WHO) criteria for diagnosing anemia (at sea level) for patients above the age of 15.}
    \label{tab:who}
\end{table}

%% file: floats/records.tex
\begin{figure}[!hp]
    \small
    \centering
    \begin{tabularx}{\linewidth}{llX | rrr | rrcr | rrcr}
    \toprule
    &&
        &\multicolumn{3}{c|}{Staging}
        &\multicolumn{4}{c|}{Mortality}
        &\multicolumn{4}{c}{Length-of-Stay} \\
    
    HADM & Timestamp 
        & Phenotypes 
        & AKI & PI & Anemia 
        & 24h & 48h & $\cdots$ & 1y 
        & 1d & 2d & $\cdots$ & 2w \\
    \midrule    
    11101 & 2012-10-04 
        & 115, 20, 305 
        & 0 & 0 & 3 
        & False & True && True 
        & False & False && True \\
    
    11101 & 2012-10-05 
        & 115,20,305, 
        & 0 & 1 & 3 
        & False & True && True 
        & False & True && True \\
        
    11102 & 2014-09-09
        & 10,140, 
        & 1 & 0 & 1 
        & False & False && False 
        & False & False && True \\
    \bottomrule
    \end{tabularx}
    \caption{Example multi-task records in CLUE.}
    \label{fig:records}
\end{figure}

%% file: floats/note.tex
\begin{figure}[!hp]
    \begin{framed}
    \obeylines%
    \small%
\hlsection{CHIEF COMPLAINT:} Abdominal pain.
\vspace{.5\baselineskip}
\hlsection{HISTORY OF PRESENT ILLNESS:} The patient is a 71-year-old female patient of Dr. X. The patient presented to the emergency room last evening with approximately 7- to 8-day history of abdominal pain which has been persistent. She was seen 3 to 4 days ago at ABC ER and underwent evaluation and discharged and had a CT scan at that time and she was told it was "normal." She was given oral antibiotics of Cipro and Flagyl. She has had no nausea and vomiting, but has had persistent associated anorexia. She is passing flatus, but had some obstipation symptoms with the last bowel movement two days ago. She denies any bright red blood per rectum and no history of recent melena. Her last colonoscopy was approximately 5 years ago with Dr. Y. She has had no definite fevers or chills and no history of jaundice. The patient denies any significant recent weight loss.
\vspace{.5\baselineskip}
\hlsection{PAST MEDICAL HISTORY:} Significant for history of atrial fibrillation, under good control and now in normal sinus rhythm and on metoprolol and also on Premarin hormone replacement.
\vspace{.5\baselineskip}
\hlsection{PAST SURGICAL HISTORY:} Significant for cholecystectomy, appendectomy, and hysterectomy. She has a long history of known grade 4 bladder prolapse and she has been seen in the past by Dr. Chip Winkel, I believe that he has not been re-consulted.
\vspace{.5\baselineskip}
\hlsection{ALLERGIES:} SHE IS ALLERGIC OR SENSITIVE TO MACRODANTIN.
\vspace{.5\baselineskip}
\hlsection{SOCIAL HISTORY:} She does not drink or smoke.
\vspace{.5\baselineskip}
\hlsection{REVIEW OF SYSTEMS:} Otherwise negative for any recent febrile illnesses, chest pains or shortness of breath.
\vspace{.5\baselineskip}
\hlsection{PHYSICAL EXAMINATION:}
\hlsubsection{GENERAL:} The patient is an elderly thin white female, very pleasant, in no acute distress.
\hlsubsection{VITAL SIGNS:} Her temperature is 98.8 and vital signs are all stable, within normal limits.
\hlsubsection{HEENT:} Head is grossly atraumatic and normocephalic. Sclerae are anicteric. The conjunctivae are non-injected.
\hlsubsection{NECK:} Supple.
\hlsubsection{CHEST:} Clear.
\vspace{.25\baselineskip}
\hlsubsection{HEART:} Regular rate and rhythm.
\hlsubsection{ABDOMEN:} Generally nondistended and soft. She is focally tender in the left lower quadrant to deep palpation with a palpable fullness or mass and focally tender, but no rebound tenderness. There is no CVA or flank tenderness, although some very minimal left flank tenderness.
\hlsubsection{PELVIC:} Currently deferred, but has history of grade 4 urinary bladder prolapse.
\hlsubsection{EXTREMITIES:} Grossly and neurovascularly intact.
\vspace{.5\baselineskip}
\hlsection{LABORATORY VALUES:} White blood cell count is 5.3, hemoglobin 12.8, and platelet count normal. Alkaline phosphatase elevated at 184. Liver function tests otherwise normal. Electrolytes normal. Glucose 134, BUN 4, and creatinine 0.7.
\vspace{.5\baselineskip}
\hlsection{DIAGNOSTIC STUDIES:} EKG shows normal sinus rhythm.
\vspace{.5\baselineskip}
\hlsection{IMPRESSION AND PLAN:} A 71-year-old female with greater than one-week history of abdominal pain now more localized to the left lower quadrant. Currently is a nonacute abdomen. The working diagnosis would be sigmoid diverticulitis. She does have a history in the distant past of sigmoid diverticulitis. I would recommend a repeat stat CT scan of the abdomen and pelvis and keep the patient nothing by mouth. The patient was seen 5 years ago by Dr. Y in Colorectal Surgery. We will consult her also for evaluation. The patient will need repeat colonoscopy in the near future and be kept nothing by mouth now empirically. The case was discussed with the patient's primary care physician, Dr. X. Again, currently there is no indication for acute surgical intervention on today's date, although the patient will need close observation and further diagnostic 
    \end{framed}
    \caption{Example Emergency Room report. To preserve privacy, this note is from MT Samples, a collection of de-identified medical transcriptions. We selected a note that is representative of the Emergency Room reports found in the MIMIC-III collection.}
    \label{fig:note}
\end{figure}